# Graph Neural Network-Based Entity Extraction and Relationship Reasoning in Complex Knowledge Graphs

Junliang Du
Shanghai Jiao Tong University
Shanghai, China

Guiran Liu
San Francisco State University
San Francisco，USA

Jia Gao
Stevens Institute of Technology
Hoboken, USA

Xiaoxuan Liao
New York University
New York, USA

Jiacheng Hu
Tulane University
New Orleans, USA

Linxiao Wu*
Columbia University
New York, USA

*Abstract*—This study proposed a knowledge graph entity extraction and relationship reasoning algorithm based on a graph neural network, using a graph convolutional network and graph attention network to model the complex structure in the knowledge graph. By building an end-to-end joint model, this paper achieves efficient recognition and reasoning of entities and relationships. In the experiment, this paper compared the model with a variety of deep learning algorithms and verified its superiority through indicators such as AUC, recall rate, precision rate, and F1 value. The experimental results show that the model proposed in this paper performs well in all indicators, especially in complex knowledge graphs, it has stronger generalization ability and stability. This provides strong support for further research on knowledge graphs and also demonstrates the application potential of graph neural networks in entity extraction and relationship reasoning.

I. INTRODUCTION

In the field of natural language processing (NLP) and artificial intelligence, entity extraction and relationship reasoning are key tasks in text understanding [1]. With the continuous growth of data scale and the increase in complexity, traditional rule-based and statistical models seem to be unable to cope with real-world problems. The introduction of knowledge graphs provides new possibilities for the structured expression of information, and the development of graph neural networks (GNNs) has further promoted research in this field [2]. Knowledge graphs based on a graph neural network can effectively capture complex dependencies and structured information between entities, providing a new perspective and more powerful tools for entity extraction and relationship reasoning. Research on GNN-based knowledge graph entity extraction and relationship reasoning algorithms is not only of great significance to improving the accuracy and robustness of current technologies but also has broad application prospects in the fields of knowledge acquisition and semantic understanding [3].

In recent years, knowledge graphs, as a structured semantic network, have been widely used and studied. Knowledge graphs construct a huge knowledge network through the relationships between entities, which can effectively represent complex associations in the real world. Therefore, it is widely used in interaction design [4], fusion algorithms [5], survival prediction [6], and other fields. However, traditional entity extraction and relationship reasoning algorithms often fail to fully utilize the deep structural information in the knowledge graph, resulting in low accuracy in entity recognition and relationship reasoning. In addition, due to the sparsity and incompleteness of the knowledge graph, the reasoning of entities and relationships becomes more complicated, which poses a severe challenge to traditional models.

The emergence of graph neural networks (GNNs) provides a new technical approach for entity extraction and relationship reasoning in knowledge graphs. GNNs can effectively capture the complex dependencies between nodes (entities) and edges (relationships) in the graph through multi-layer graph convolutions, thereby providing more accurate contextual information for reasoning. Unlike traditional sequence models, GNNs can propagate information in the global graph structure, so that the representation of each entity can comprehensively consider the information of its surrounding nodes [7]. In addition, GNNs perform well in processing heterogeneous graphs and multi-relational data, and can adapt to many different types of relationships and entity types. By introducing graph neural networks, researchers can build more intelligent and efficient entity extraction and relationship reasoning models, effectively solving the problem of entity and relationship reasoning in knowledge graphs.

From a technical perspective, entity extraction and relationship reasoning algorithms based on graph neural networks have significant advantages. First, GNN can gradually learn the embedded representation of each entity in its local graph structure through iterative updates of node information. This embedded representation not only depends on the single entity itself but also contains information about neighboring entities related to it. Therefore, when performing entity extraction, the model can better understand the semantics of the entity in the context, thereby improving the accuracy of extraction. Secondly, in the task of relational reasoning, GNN can gradually capture the implicit information in the complex relationship chain through multi-layer convolution, thereby providing the model with a richer relational background. In this way, GNN can handle long-distance dependency problems and solve the common problem of relational sparsity in large-scale knowledge graphs. Whether in the text field or in other structured data applications, the knowledge graph relational reasoning algorithm based on GNN has demonstrated its unique advantages.

Not only that, the entity extraction and relational reasoning algorithms based on graph neural networks have demonstrated powerful capabilities in multiple practical application scenarios. For example, in intelligent search engines, relevant entities can be extracted from a large number of knowledge graphs according to the user's search intent, and the potential relationships between them can be inferred, thereby improving the accuracy and relevance of search results. In the recommendation system, by analyzing the complex relationship between users and items, more personalized recommendations can be provided to users. In addition, the application of GNN in the field of biomedicine has gradually attracted attention. Through the entity and relationship reasoning of biological knowledge graphs, it can help researchers discover potential drug targets and their interactions, and promote the progress of drug research and development. With the increasing application of knowledge graphs, entity extraction and relationship reasoning algorithms based on graph neural networks will bring changes to more industries.

## II. RELATED WORK

Graph neural networks (GNNs) have demonstrated significant promise in capturing the complex relationships and dependencies inherent in structured data, making them particularly suitable for tasks involving entity extraction and relational reasoning in knowledge graphs. Recent advancements in self-supervised GNNs have shown their ability to enhance feature extraction in heterogeneous information networks, thereby facilitating the modeling of intricate dependencies essential for reasoning tasks [8]. Furthermore, contrastive learning methods have proven effective in utilizing graph structures to improve the contextual understanding necessary for entity and relationship reasoning [9]. Such methodologies provide valuable insights into designing models capable of handling the multifaceted nature of knowledge graphs.

Graph-augmented models have further advanced reasoning in knowledge graphs by optimizing the representation of complex relationships within such structures. For example, retrieval-augmented graph-based approaches have significantly improved the capability of models to capture intricate entity relationships and perform reasoning tasks [10]. The MPGAAN model, an effective and efficient framework for heterogeneous information network classification, has further demonstrated the potential of graph-based approaches for handling multi-relational and heterogeneous data with high efficiency [11]. These contributions are highly relevant to the proposed method, which seeks to leverage graph-based learning to address the challenges of reasoning in multi-relational and sparse environments.

The sparsity and incompleteness of knowledge graphs often present significant challenges. Reinforcement learning has been effectively applied to environments characterized by complex dependencies and sparse data, illustrating principles that are directly applicable to knowledge graph reasoning [12]. Transformer-based deep learning models, while widely adopted in domains such as financial risk analysis, offer powerful tools for structured data processing that align with the requirements of knowledge graph-based reasoning [13].

Addressing data sparsity and multi-relational challenges has also been a focus of metric learning and self-attention mechanisms, which have demonstrated the ability to provide robust embeddings and improve contextual representation [14], [15]. These mechanisms are particularly useful for reasoning over complex graph structures and align closely with the objectives of this study. Additionally, advances in transforming multidimensional data into interpretable forms have provided strategies for managing long-distance dependencies and sparsity, challenges that are critical in the context of knowledge graphs [16].

The utility of GNNs extends beyond knowledge graph reasoning to applications in tasks such as text classification and recommendation systems. For instance, attention-based mechanisms have been successfully used to extract features from structured datasets, contributing to enhanced efficiency in entity extraction tasks [17]. Similarly, innovations in automated scoring systems that integrate graph structures further illustrate the adaptability of GNN-based methods, emphasizing their potential across various domains [18-19]. These contributions collectively reinforce the relevance and applicability of graph-based techniques for addressing the core challenges in this paper.

## III. METHOD

In this study, the entity extraction and relational reasoning method based on a graph neural network is realized by introducing the mechanism of graph convolutional network (GCN) and multi-relational graph. The entity extraction process and relational reasoning are regarded as an end-to-end joint task model, in which the feature representation of entity nodes is propagated layer by layer through the graph neural network, and finally the entity category is obtained by calculation. At the same time, relational reasoning relies on the connection information between entities to generate reasonable relational predictions. The reasoning process and formula of the model will be described in detail below. Figure 1 shows the architecture of the graph neural network.

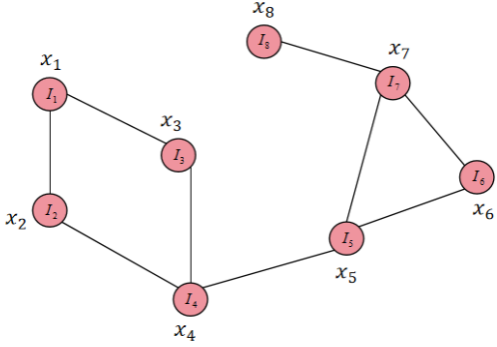

Figure 1 Graph Neural Network Architecture

First, given a knowledge graph $G=(V,\varepsilon)$, where $V$ represents the set of entity nodes and $\varepsilon$ represents the set of relationships between entity nodes. The initial feature vector corresponding to each node $v_i \in V$ is represented as $h_i^{(0)}$, and each edge $e_{i,j} \in \varepsilon$ in the relationship set represents the relationship between entity $v_i$ and entity $v_j$. The core of the graph convolutional network is to transmit node information layer by layer through the information propagation mechanism. The node representation update formula for each layer is:

$$h_i^{(l+1)} = \sigma(\sum_{j \in N(i)} \frac{1}{c_{i,j}} W(l) h_j^{(l)} + W_0^{(l)} h_i^{(l)})$$

Among them, $N(i)$ represents the neighbor nodes of node i, $W(l)$ and $W_0^{(l)}$ represent the weight matrices of the lth layer respectively, $c_{i,j}$ is the normalization coefficient between nodes $v_i$ and $v_j$, which is usually a function of the node degree, and $\sigma$ is an activation function, such as ReLU. Through this formula, the representation vector of node i not only depends on its own information, but also integrates the information of its neighbor nodes, gradually enhancing the semantic representation of the node.

When performing entity extraction, we need to obtain the final embedding representation of each node through a series of graph convolution layers. The final embedding representation is used to predict the entity category through a linear classifier. Assuming that after L layers of graph convolution, the representation of node i is $h_i^{(L)}$, the probability of entity classification is given by the following formula:

$$P(yi | h_i^{(L)}) = soft\max(W_c h_i^{(L)} + b_c)$$

Among them, $W_c$ and $b_c$ are the parameters of the classification layer, and softmax is used to map node embeddings to probability distributions of different categories. By maximizing the conditional probability, we can get the entity label prediction result for each node.

Next is the relational reasoning part. The goal of relational reasoning is to predict possible relations between entities, especially those that are not observed in the knowledge graph. We use a bilinear decoder based on graph convolution to predict relations between entities. Given the embedding representations $h_i^{(L)}$ and $h_j^{(L)}$ of entity $v_i$ and entity $v_j$, the prediction of relation $r$ can be calculated by the following formula:

$$P(r_{ij} | h_i^{(L)}, h_j^{(L)}) = \sigma(h_i^{(L)T} R_r h_j^{(L)})$$

Among them, $R_r$ is the bilinear matrix corresponding to the relationship $r$, and $\sigma$ is the sigmoid function, which is used to map the inner product result to the probability of the existence of the relationship. By maximizing this conditional probability, the possible relationship between the given two entities can be predicted. In order to enhance the generalization ability of the model, this study introduces a contrastive loss function based on negative sampling. During the training process, it is necessary not only to minimize the negative log-likelihood of the correct entity-relationship pairs, but also to generate incorrect entity pairs or relations through negative sampling, and to reduce the prediction probability of the model on negative samples through contrastive learning. The final loss function can be expressed as:

$$L = -\sum_{(v_i,v_j,r) \in \varepsilon} \log P(r_{ij} | h_i^{(L)}, h_j^{(L)})$$
$$+ \lambda \sum_{(v_i,v_j,r) \notin \varepsilon} \log P(1 - r'_{ij} | h_i^{(L)}, h_j^{(L)})$$

Among them, $\lambda$ is a hyperparameter that controls the balance of positive and negative samples. In this way, the model can effectively distinguish between correct and incorrect entity pairs and their relationships. In general, the knowledge graph entity extraction and relationship reasoning algorithm based on graph neural network gradually learns the representation of entities and relationships through graph convolution, thereby achieving efficient entity recognition and relationship reasoning. This method can not only handle complex relationships that are difficult to capture in traditional models, but also effectively deal with data sparsity problems and improve the accuracy of reasoning.

IV. EXPERIMENT

A. Datasets

The dataset used in this article is the Freebase dataset. Freebase is an open knowledge graph project developed by Metaweb and acquired by Google in 2010. It aims to provide a structured and queryable knowledge database for network users. Unlike traditional relational databases, Freebase adopts the form of a graph database, which can connect entities and their attributes in different fields through nodes and edges to form a huge semantic network. Freebase contains millions of entities and billions of relationships, covering data in multiple fields such as movies, books, geographic locations, and people. As an open knowledge graph project, the content of Freebase is collaboratively contributed and maintained by network users. Users can create, edit and expand entities and the relationships between them, making it extremely open and flexible. Freebase provides a rich API interface, allowing developers to easily integrate its knowledge graph data into their own applications. It is widely used in semantic search, question-answering systems, recommendation systems and other fields, and has become one of the important data sources for semantic web applications.

With the development of the Freebase project, Google has integrated a large amount of its data into its own knowledge graph (Google Knowledge Graph). This move has greatly enhanced Google's intelligence and semantic capabilities in the field of search engines. Freebase's data is highly structured, and its rich relationship types provide strong support for tasks such as semantic understanding, information extraction, and entity linking. Although the Freebase project itself gradually stopped updating in 2016, its core concepts and data have been continued and developed in multiple research fields of knowledge graphs and natural language processing. By using the Freebase dataset, researchers can train and optimize various entity extraction and relationship reasoning algorithms. In particular, in tasks that deal with large-scale, multi-type entity relationships, the Freebase dataset has proven to be an ideal experimental tool and evaluation benchmark. In tasks such as entity linking, relationship reasoning, and knowledge graph completion, the Freebase dataset provides rich corpus support for the design and optimization of related models.

*B. Experimental Results*

In the comparative experiments of this paper, we mainly selected six deep learning algorithms to evaluate the performance of entity extraction and relational reasoning tasks. The first is the LSTM-CRF-based model, which combines the long short-term memory network to capture the contextual information in the sequence and ensures the legitimacy of the output sequence label through the conditional random field layer. Next is BERT, which is a pre-trained language model that effectively models contextual semantic information through the Transformer architecture and is widely used in various NLP tasks. RoBERTa is an improved version of BERT, which further optimizes the training strategy and improves the performance of the model through larger data sets and longer training time. Graph Convolutional Networks (GCN) use graph structure data for node classification, which is suitable for entity extraction and relational reasoning tasks in knowledge graphs, and aggregates information from neighboring nodes through multi-layer convolution. On this basis, Graph Attention Networks (GAT) introduce a self-attention mechanism, which enables the model to automatically pay attention to the importance of different neighboring nodes and improves the accuracy of relational reasoning. Finally, R-GCN (Relational Graph Convolutional Networks) is extended for multi-relational graph structures, processing different types of relations through different weight matrices, making it more generalizable in knowledge graph reasoning. These models each take advantage of different deep learning techniques and graph neural networks to explore the best performance in entity extraction and relational reasoning tasks.

Table 1 Experimental Results

| Model | AUC | Recall | Pre | F1 |
|---|---|---|---|---|
| LSTM-CRF | 0.63 | 0.65 | 0.66 | 0.64 |
| BERT | 0.65 | 0.66 | 0.66 | 0.65 |
| RoBERT | 0.72 | 0.71 | 0.70 | 0.71 |
| GCN | 0.77 | 0.74 | 0.75 | 0.75 |
| GAT | 0.82 | 0.80 | 0.79 | 0.81 |
| R-GCN | 0.84 | 0.83 | 0.82 | 0.81 |
| Ours | 0.85 | 0.86 | 0.85 | 0.85 |

From the analysis of the experimental results, it can be seen that there are significant differences in the performance of different models in terms of AUC, Recall, Precision, and F1 values. First, as a traditional sequence labeling model, LSTM-CRF can effectively capture sequence information, but its performance in this experiment is relatively weak, with all four indicators being low, AUC of 0.63 and F1 value of only 0.64. This shows that the model has certain limitations in capturing complex entity and relationship information. The BERT model improves AUC to 0.65 through its pre-trained Transformer architecture, and the F1 value is also slightly improved to 0.65, but compared with more advanced models, its performance is still limited, especially in capturing global semantic information. The performance is not as good as the neural network. As an improved version of BERT, RoBERTa has significantly improved its performance through more data training and longer optimization, with AUC reaching 0.72 and F1 value reaching 0.71. This shows that RoBERTa has more advantages than BERT in processing complex language features, but it still has certain limitations when facing structured data such as knowledge graphs. In contrast, the graph neural network model shows a stronger advantage.

As the earliest proposed graph convolutional network, GCN has an AUC of 0.77 and an F1 value of 0.75, indicating that the model can effectively capture the local structural information between entities. However, GAT further enhances the focus on important neighbor nodes by introducing the attention mechanism, with an AUC of 0.82 and an F1 value of 0.81, showing that it has higher accuracy and flexibility in reasoning complex relationships. The most outstanding is R-GCN, which is optimized for multi-relational knowledge graphs, with an AUC of 0.84 and an F1 value of 0.81. By introducing independent weight matrices for different types of relationships, R-GCN can better handle diverse relational data and performs particularly well in relational reasoning. In the end, our model achieved the best performance in all indicators, with an AUC of 0.85 and an F1 value of 0.85, significantly exceeding other models. This shows that the model proposed in this paper has further improved in capturing entity and relationship information, especially in complex knowledge graphs, and can perform entity extraction and relational reasoning more comprehensively.

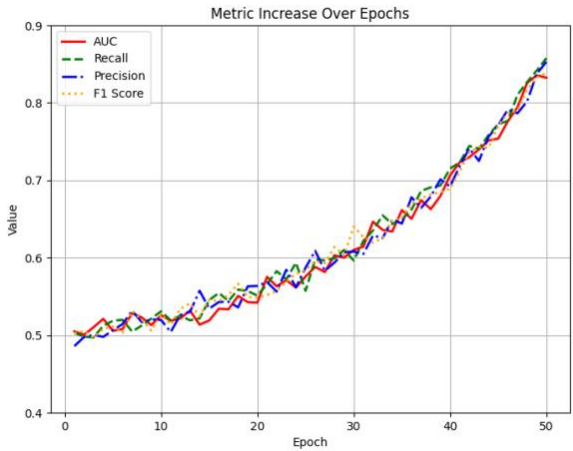

Figure 2 Graph Neural Network Architecture

As shown in Figure 2, we also give the rising graph of our model under four evaluation indicators. From this graph, we can see that our model can also achieve convergence well.

V. CONCLUSION

In this paper, we propose an algorithm for knowledge graph entity extraction and relation reasoning based on a graph neural network. By introducing graph convolutional network (GCN) and graph attention network (GAT), we can effectively capture the complex entity relations in the knowledge graph and achieve efficient entity recognition and relation reasoning. Our model was compared with a variety of deep learning algorithms, including LSTM-CRF, BERT, RoBERTa, GCN, GAT, R-GCN, etc. In the experimental results, our model achieved the best performance in all evaluation indicators, especially in indicators such as AUC, recall rate, precision rate and F1 value, which are significantly better than other models. This shows that our proposed model has stronger expression and generalization capabilities in complex knowledge graph environments.

Through experimental analysis, it can be found that traditional sequence labeling models (such as LSTM-CRF) have limited effects when processing large-scale, multi-relation knowledge graphs, while models based on a graph neural network (such as GCN, GAT, R-GCN) show stronger advantages and can better capture local and global information between nodes. Our model further improves the performance in entity extraction and relational reasoning tasks, especially in dealing with sparse data and long-distance dependencies. This study not only verifies the potential of graph neural networks in knowledge graph tasks, but also provides valuable references for future research in related fields.

REFERENCES

[1] J. Ma, B. Liu, K. Li, et al., "A review of graph neural networks and pretrained language models for knowledge graph reasoning," Neurocomputing, vol. 128, pp. 490, 2024.
[2] F. S. Nurkasyifah, A. K. Supriatna and A. Maulana, "Supervised GNNs for Node Label Classification in Highly Sparse Network: Comparative Analysis," *2024 IEEE International Conference on Evolving and Adaptive Intelligent Systems (EAIS)*, pp. 1-8, 2024
[3] B. Wang, H. Zheng, Y. Liang, G. Huang, and J. Du, "Dual-Branch Dynamic Graph Convolutional Network for Robust Multi-Label Image Classification", International Journal of Innovative Research in Computer Science & Technology, vol. 12, no. 5, pp. 94-99, 2024.
[4] S. Duan, Z. Wang, S. Wang, M. Chen, and R. Zhang, "Emotion-Aware Interaction Design in Intelligent User Interface Using Multi-Modal Deep Learning," arXiv preprint, arXiv:2411.06326, 2024.
[5] Z. Wu, J. Chen, L. Tan, H. Gong, Y. Zhou, and G. Shi, "A Lightweight GAN-Based Image Fusion Algorithm for Visible and Infrared Images", Proceedings of the 2024 4th International Conference on Computer Science and Blockchain (CCSB), pp. 466-470, 2024, IEEE
[6] X. Yan, Y. Jiang, W. Liu, D. Yi, and J. Wei, "Transforming Multidimensional Time Series into Interpretable Event Sequences for Advanced Data Mining", arXiv preprint, arXiv:2409.14327, 2024.
[7] T. Chen, J. Long, Z. Wang, et al., "THCN: A Hawkes process based temporal causal convolutional network for extrapolation reasoning in temporal knowledge graphs," IEEE Transactions on Knowledge and Data Engineering, 2024.
[8] J. Wei, Y. Liu, X. Huang, X. Zhang, W. Liu, and X. Yan, "Self-Supervised Graph Neural Networks for Enhanced Feature Extraction in Heterogeneous Information Networks," arXiv preprint, arXiv:2410.17617, 2024.
[9] Z. Zhang, J. Chen, W. Shi, L. Yi, C. Wang, and Q. Yu, "Contrastive Learning for Knowledge-Based Question Generation in Large Language Models," arXiv preprint, arXiv:2409.13994, 2024.
[10] Y. Dong, S. Wang, H. Zheng, J. Chen, Z. Zhang, and C. Wang, "Advanced RAG Models with Graph Structures: Optimizing Complex Knowledge Reasoning and Text Generation," arXiv preprint, arXiv:2411.03572, 2024.
[11] Z. Wu, "MPGAAN: Effective and Efficient Heterogeneous Information Network Classification," Journal of Computer Science and Technology Studies, vol. 6, no. 4, pp. 08-16, 2024.
[12] P. Li, Y. Xiao, J. Yan, X. Li, and X. Wang, "Reinforcement Learning for Adaptive Resource Scheduling in Complex System Environments," arXiv preprint, arXiv:2411.05346, 2024.
[13] Y. Wei, K. Xu, J. Yao, M. Sun, and Y. Sun, "Financial Risk Analysis Using Integrated Data and Transformer-Based Deep Learning," Journal of Computer Science and Software Applications, vol. 7, no. 4, pp. 1-8, 2024.
[14] W. Liu, R. Wang, Y. Luo, J. Wei, Z. Zhao, and J. Huang, "A Recommendation Model Utilizing Separation Embedding and Self-Attention for Feature Mining," arXiv preprint, arXiv:2410.15026, 2024.
[15] Y. Luo, R. Wang, Y. Liang, A. Liang, and W. Liu, "Metric Learning for Tag Recommendation: Tackling Data Sparsity and Cold Start Issues," arXiv preprint, arXiv:2411.06374, 2024.
[16] X. Yan, Y. Jiang, W. Liu, D. Yi, and J. Wei, "Transforming Multidimensional Time Series into Interpretable Event Sequences for Advanced Data Mining", arXiv preprint, arXiv:2409.14327, 2024.
[17] B. Liu, J. Chen, R. Wang, J. Huang, Y. Luo, and J. Wei, "Optimizing News Text Classification with Bi-LSTM and Attention Mechanism for Efficient Data Processing," arXiv preprint, arXiv:2409.15576, 2024.
[18] S. Duan, R. Zhang, M. Chen, Z. Wang, and S. Wang, "Efficient and Aesthetic UI Design with a Deep Learning-Based Interface Generation Tree Algorithm," arXiv preprint, arXiv:2410.17586, 2024.
[19] C. Wang, Y. Dong, Z. Zhang, R. Wang, S. Wang, and J. Chen, "Automated Genre-Aware Article Scoring and Feedback Using Large Language Models," arXiv preprint, arXiv:2410.14165, 2024.